\documentclass{bmvc2k}

\usepackage{graphicx}
\usepackage{amsmath}
\usepackage{amssymb}
\usepackage{booktabs}
\usepackage[skip=5pt]{caption}
\usepackage[inline]{enumitem}
\usepackage{placeins}

\title{Vision Backbone Efficient Selection for Image Classification in Low-Data Regimes}

\addauthor{Joris Guerin}{joris.guerin@ird.fr}{1}
\addauthor{Shray Bansal}{sbansal34@gatech.edu}{2}
\addauthor{Amirreza Shaban}{amir@fieldai.com}{3}
\addauthor{Paulo Mann}{paulomann@dcc.ufrj.br}{4}
\addauthor{Harshvardhan Gazula}{hvgazula@mit.edu}{5}

\addinstitution{Espace-Dev, IRD, University \\ of Montpellier}
\addinstitution{Georgia Institute of Technology}
\addinstitution{Field AI}
\addinstitution{Institute of Computing, Federal \\ University of Rio de Janeiro}
\addinstitution{Massachusetts Institute of \\ Technology}

\runninghead{Guerin et al.}{VIBES}

\begin{document}

\maketitle

\begin{abstract}
%
Transfer learning has become an essential tool in modern computer vision, allowing practitioners to leverage backbones, pretrained on large datasets, to train successful models from limited annotated data. Choosing the right backbone is crucial, especially for small datasets, since final performance depends heavily on the quality of the initial feature representations. While prior work has conducted benchmarks across various datasets to identify universal top-performing backbones, we demonstrate that backbone effectiveness is highly dataset-dependent, especially in low-data scenarios where no single backbone consistently excels. To overcome this limitation, we introduce dataset-specific backbone selection as a new research direction and investigate its practical viability in low-data regimes. Since exhaustive evaluation is computationally impractical for large backbone pools, we formalize Vision Backbone Efficient Selection (VIBES) as the problem of searching for high-performing backbones under computational constraints. We define the solution space, propose several heuristics, and demonstrate VIBES feasibility for low-data image classification by performing experiments on four diverse datasets. Our results show that even simple search strategies can find well-suited backbones within a pool of over $1300$ pretrained models, outperforming generic benchmark recommendations within just ten minutes of search time on a single GPU (NVIDIA RTX A5000) 
\end{abstract}

\section{Introduction}
\label{sec:introduction}

Transfer learning is crucial in low-data computer vision (CV) scenarios, where practitioners must extract maximum value from limited annotated examples. Developers frequently default to well-established pretrained backbones like ResNet~\cite{resnet} or Vision Transformers (ViT)~\cite{vit} without thoroughly assessing their suitability for the specific dataset at hand. This oversight is significant, as recent studies have demonstrated that different pretrained backbones can exhibit vastly different generalization capabilities across downstream domains~\cite{guerin2021combining, gontijo-lopes2022no}. This is particularly problematic in low-data regimes, where initial feature quality determines final performance, as the model has minimal opportunity to adapt during fine-tuning. In this work, we argue that a more deliberate approach to backbone selection can yield significant performance gains with minimal additional effort compared to hyperparameter tuning or architectural modifications.

\begin{figure}[t]
    \centering
    \includegraphics[width=0.95\textwidth]{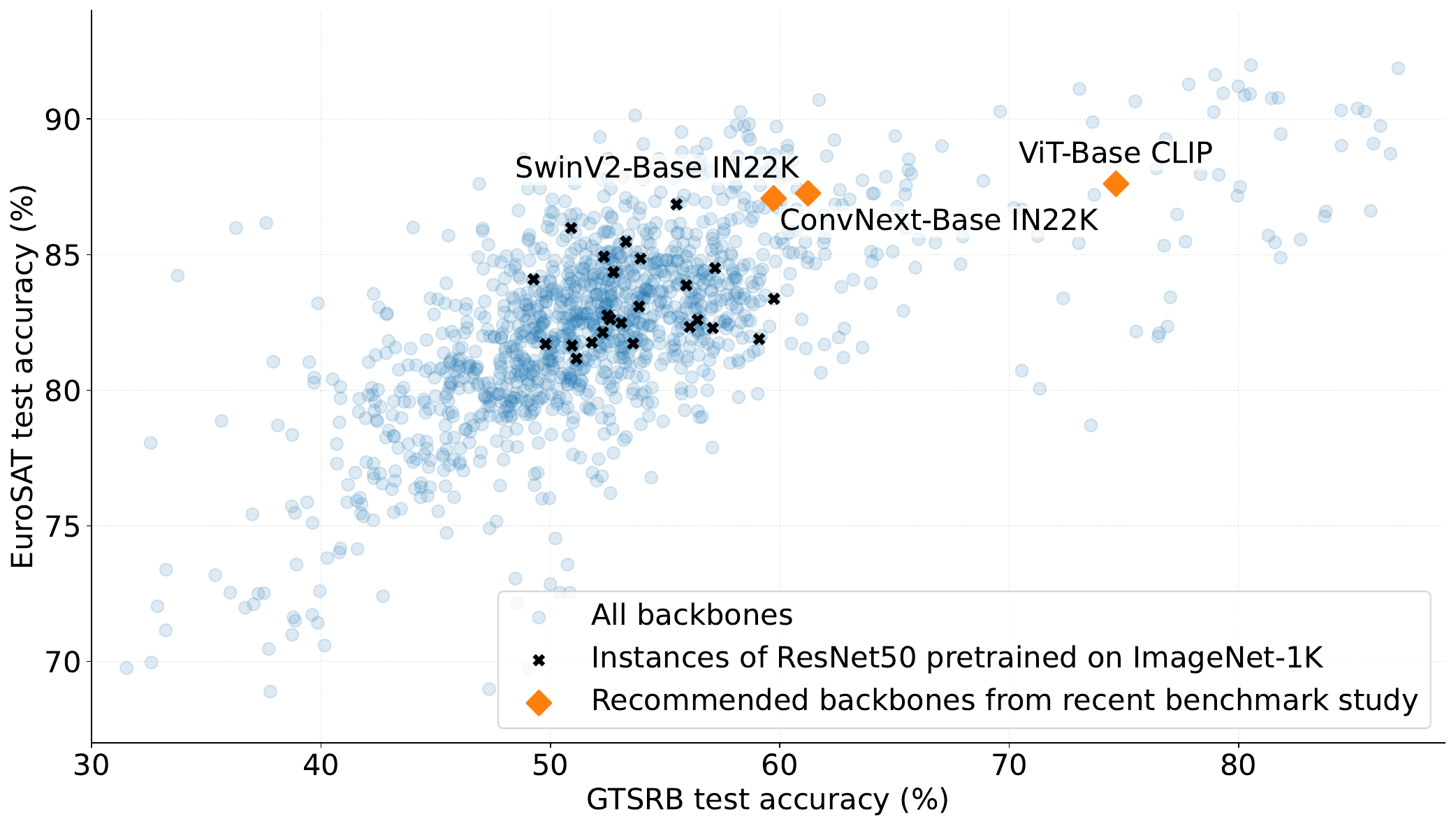}
    \caption{\textbf{Backbone performance across 2 datasets in low-data regimes}. Each point represents a pretrained backbone evaluated by training a logistic regressor on frozen features using only 10 samples per class. Recommended backbones are from~\cite{goldblum2024battle}.}
    \label{fig:issues_benchmarks}
\end{figure}

To date, the topic of vision backbone selection has primarily been addressed through benchmark studies (Section~\ref{sec:related_work}), which evaluate a range of pretrained backbones across multiple downstream datasets. By aggregating performances, these studies aim to propose general recommendations, such as identifying overall top-performing architectures or noting trends in the effectiveness of certain backbone families. 
While these studies provide valuable insights, they face fundamental limitations:
\begin{itemize}[leftmargin=0pt, label={}, itemsep=3pt, parsep=0pt, topsep=2pt]

\item \textbf{Limited coverage:} Benchmarks struggle to encompass the vast array of available backbones. Recent studies compared fewer than 50 models, while deep learning model libraries often offer over 1000 options. Moreover, the rapidly evolving landscape of CV models means that benchmark results can quickly become outdated as new architectures emerge.

\item \textbf{Overemphasis on average performance:} Benchmarks inherently focus on average-case performance, overlooking significant variations across real-world datasets. As shown in Figure~\ref{fig:issues_benchmarks}, the three generic recommendations from the most recent benchmark study~\cite{goldblum2024battle} display suboptimal behavior on both datasets. 

\item \textbf{Neglecting Implementation Variability:} Benchmarks provide results based on high-level model descriptions (e.g., architecture, pretraining method). However, they fail to capture the significant performance variability among models with identical specifications. Figure~\ref{fig:issues_benchmarks} shows multiple ResNet50 models, all trained using supervised learning on ImageNet-1K, exhibiting significantly different performance across both datasets. This implementation-specific variability is inherently difficult for benchmark studies to capture. Practitioners relying solely on benchmark results may overlook optimal model variants for their tasks.
\end{itemize}
To address these limitations, \emph{we propose dataset-specific backbone selection as a novel research direction}. The goal is to identify the most suitable backbone for a given dataset from a set of available pretrained models. This rather simple formulation represents a fundamental departure from traditional practices by: (1) dynamically considering a vast array of backbones, including newly released models, (2) prioritizing dataset-specific performance over average-case scenarios, (3) accounting for implementation-specific variability by evaluating individual model instances.

While exhaustive search guarantees optimal backbone selection, it quickly becomes intractable for large sets of pretrained backbones. Hence, we formalize the Vision Backbone Efficient Selection (VIBES) problem (Section \ref{sec:problem}), which aims to identify the best backbone under computational constraints, trading off between optimality and efficiency. To demonstrate the feasibility of dataset-specific backbone selection, we propose several search heuristics (Section \ref{sec:approach}) and evaluate them across four diverse low-data image classification datasets (Section \ref{sec:experiments}). Our results demonstrate that even simple dataset-specific backbone selection strategies like random search can outperform generic benchmarks within just ten minutes on a single GPU (NVIDIA RTX A5000 24GB). These findings establish dataset-specific backbone selection as an important new research direction that could fundamentally change how practitioners approach transfer learning in practical CV applications, particularly in low-data scenarios where labeled examples are scarce and expensive to produce.

\section{Related work: the elusive quest for a universal backbone}
\label{sec:related_work}

Transfer learning consists of using pretrained neural network backbones as  feature extractors for smaller-scale downstream tasks~\cite{sharif2014cnn}. However, the number of available pretrained backbones has grown exponentially, due to the multiplication of novel architectures, training algorithms, and pretraining datasets. Indeed, the \textit{timm} library~\cite{rw2019timm} now hosts over 1300 pretrained backbones, highlighting the breadth of options available. Here, we survey the literature aimed at guiding practitioners through this extensive array of backbones. 

Kornblith et al.~\cite{kornblith2019better} studied the correlation between performance on ImageNet and other datasets and proposed ImageNet accuracy as a proxy for downstream backbone performance. However, subsequent research challenged this assumption, demonstrating that good performance on ImageNet does not always translate to enhanced efficacy on real-world datasets~\cite{fang2024does}.

Beyond these correlation studies, various benchmarks were also conducted, comparing the influence of specific backbone characteristics across a wide range of datasets to draw general conclusions about which models are likely to perform well on new datasets. 
Goldblum et al.~\cite{goldblum2024battle} conducted the most extensive benchmark study to date, comparing over 20 backbones along three axes: architecture, pretraining algorithm, and pretraining dataset. Their comprehensive analysis across many downstream datasets provides broad conclusions regarding optimal backbones, along with targeted recommendations for specific families of tasks.
Vishniakov et al.~\cite{vishniakov2024convnet} compared two architectures (ConvNeXt~\cite{convnext} and ViT~\cite{vit}) across two training methodologies (supervised and CLIP~\cite{clip}). Their findings suggest that the optimal choice depends on target data attributes, reinforcing our hypothesis that universally optimal recommendations are unlikely to exist.
Jeevan et al.~\cite{jeevan2024backbone} focused on benchmarking lightweight convolutional architectures under consistent training settings.
Ericsson et al.~\cite{ericsson2021well} examined self-supervised models, comparing 13 backbones across over 40 datasets, concluding that identifying a method that consistently outperforms others on downstream datasets remains challenging.
Zhai et al.~\cite{zhai2019large} constructed a pool of 19 benchmark datasets aimed at better representing the diversity of downstream domains encountered by practitioners, and used it to compare 18 supervised and self-supervised pretraining algorithms.
Kolesnikov et al.~\cite{kolesnikov2019revisiting} evaluated the transfer performance of four pretext tasks on three CNN architectures.
Finally, Goyal et al.~\cite{goyal2019scaling} investigated the influence of dataset size on self-supervised pretraining. 

While benchmark studies provide valuable insights, their generic conclusions often fall short in addressing the specific requirements of downstream domains.  
In response, this work introduces the problem of finding a backbone specifically tailored to the target dataset. 

\vspace{-10pt}
\section{Problem formulation}
\label{sec:problem}
\vspace{-5pt}

We present a general formulation of dataset-specific backbone selection that could accommodate various CV tasks, though our analysis focuses specifically on image classification in low-data settings, where the benefits are most pronounced. This formulation serves both our immediate investigation and provides a framework for future extensions to other domains.

\vspace{-10pt}
\subsection{Vision backbone selection}
\label{sec:vbs}
Let $\mathcal{T}$ denote the target CV task, characterized by a training dataset $\mathcal{D}_\text{train}$, a validation dataset $\mathcal{D}_\text{val}$, a test dataset $\mathcal{D}_\text{test}$, and an evaluation metric $\epsilon$. For a model $m$, we note $\epsilon_{\text{test}}(m)$ its performance on $\mathcal{D}_\text{test}$ with respect to $\epsilon$. Likewise, $\epsilon_{\text{val}}(m)$ denotes the validation performance. For image classification, $\epsilon_{\text{test}}(m)$ is the test accuracy of $m$, trained on $\mathcal{D}_\text{train}$. The low-data regime is characterized by small $\mathcal{D}_\text{train}$ and $\mathcal{D}_\text{val}$ (e.g., 10 samples per class).

We introduce vision backbone selection as the problem of finding the optimal backbone for~$\mathcal{T}$. Let $\mathcal{B} = \{b_1, ..., b_N\}$ be a set of $N$ pretrained backbones. Suppose we have a procedure to train a model on top of a given backbone $b\in\mathcal{B}$ for $\mathcal{T}$. For example, in our low-data classification context, this procedure involves training a simple classifier (e.g., logistic regression) on features extracted from $\mathcal{D}_\text{train}$ using $b$. We deliberately use simple classifiers with minimal parameters to avoid overfitting, which is particularly problematic when training data is scarce. The training methodology (model architecture and hyperparameters) remains constant throughout the backbone selection process to ensure that differences in performance can be attributed to the choice of the backbone. Under this assumption, we can simplify notations and use $\epsilon_{\text{test}}(b)$ and $\epsilon_{\text{val}}(b)$ to denote the performance of a backbone $b$ at task $\mathcal{T}$. 

With the above notations, vision backbone selection can be formulated as the problem of finding $b^*_{\text{test}} = {\arg\max}_{b \in \mathcal{B}} \, \epsilon_{\text{test}}(b)$. In machine learning terminology, this is a parameter selection problem in a discrete set. To avoid overfitting to the test set, we cannot evaluate models directly on $\mathcal{D}_\text{test}$ during backbone selection. Instead, we must use $\epsilon_{\text{val}}(b)$ as a proxy for $\epsilon_{\text{test}}(b)$. When $\mathcal{D}_\text{val}$ and $\mathcal{D}_\text{test}$ are both representative of true underlying data distribution of the problem, $b^*_{\text{val}} = {\arg\max}_{b \in \mathcal{B}} \, \epsilon_{\text{val}}(b)$ should closely approximate the optimal backbone. This proxy-based selection is particularly important in low-data regimes, where the risk of overfitting increases due to the small train and validation set sizes.

Since $\mathcal{B}$ is finite, $b^*_{\text{val}}$ can be found through exhaustive search. This approach runs in a total time of $t=\sum_{b\in\mathcal{B}} \tau(b)$,
where $\tau(b)$ is the time needed to extract features from $\mathcal{D}_\text{train}$ using $b$, train the classifier on these features, and evaluate its performance on $\mathcal{D}_\text{val}$. In practice, $\tau(b)$ also often includes the time to download pretrained weights.
However, exhaustive search quickly becomes impractical for large backbone pools, as training and evaluating all the models requires considerable time and computational resources, which are often limited in the scenarios where training data is scarce. While this level of resource commitment might be feasible for critical tasks in well-funded organizations, it is not a scalable solution as deep learning continues to expand across various domains.


\vspace{10pt}
\subsection{Vision Backbone Efficient Selection (VIBES)}
\label{subs:strats}

To address the above time and resource 
constraints, we introduce Vision Backbone Efficient Selection (VIBES), a relaxation of the original problem that aims to quickly find a high-performing backbone, potentially trading off optimality for speed.

To reduce $t$, we have two mathematical options: reduce $\tau(b)$ or reduce $|\mathcal{B}|$. This leads to two families of strategies to reduce the total search time:
\begin{enumerate}[leftmargin=*, itemsep=5pt, parsep=0pt]
    \item \textbf{Fast approximate evaluation}: One approach is to reduce $\tau(b)$ by defining a fast alternative evaluation procedure $\tilde{\epsilon}$, such that $\tilde{\epsilon}_{\text{val}}(b)$ is an approximation of $\epsilon_{\text{val}}(b)$. Although approximate evaluation could lead to suboptimal choices, it should be fast to compute ($\forall b \in \mathcal{B}, \tilde{\tau}(b) < \tau(b)$), resulting in a quicker exploration of the search space.

    \item \textbf{Optimized sampling}: Another approach is to use only a subset of $\mathcal{B}$. While this could mean missing out on the best backbone, it can significantly reduce search time. The quality of the selected backbone depends heavily on the order in which models are sampled. To manage this, we define a (potentially stochastic) sampling function $\pi$, which generates a permutation of $\{1, ..., N\}$. We denote the ordered set as $\pi(\mathcal{B}) = \{b_{\pi(1)}, \dots, b_{\pi(N)}\}$.

\end{enumerate}
Providing a solution to VIBES consists of defining an approximate evaluation procedure $\tilde{\epsilon}$ and
a sampling strategy $\pi$. Then, we can use $\tilde{\epsilon}$ to evaluate multiple backbones, sampled with $\pi$, for a predefined time budget $t_{\text{max}}$. The selected backbone $\hat{b}$ is defined as
\begin{equation}
    \hat{b} = \underset{i \in \{1, ..., k\}}{\arg\max} \, \tilde{\epsilon}_{\text{val}}(b_{\pi(i)}),
\end{equation}
where $k$ is the largest integer in $\{1, ..., N\}$ such that $\sum_{i=1}^{k} \tilde{\tau}(b_{\pi(i)}) \leq t_{\text{max}}$.

\subsection{Evaluation}\label{sec:problem_bsec}

The performance of a strategy $(\pi, \tilde{\epsilon})$ is measured with the true evaluation metric for the selected model: $\epsilon_{\text{test}}(\hat{b})$. This evaluation depends on the allocated time budget, and two strategies can only be compared for a given $t_{\text{max}}$. Some strategies might be better suited for short time budget while other are more performant for long searches. To rigorously assess and compare VIBES strategies, we introduce Backbone Selection Efficiency Curves (BSEC). 

Constructing a BSEC consist of plotting $\epsilon_{\text{test}}(\hat{b})$ as a function of $t_{\text{max}}$. To account for the stochastic nature of VIBES algorithms, we run each strategy multiple times and plot both a line representing a measure of central tendency, and a representation of the performance variability. In this work, we use the median and the 25th--75th percentile interval, based on 30 runs per strategy. This choice is robust to outliers and provides a clear picture of the typical performance and its spread. However, researchers may opt for other measures (e.g., mean and standard deviation), if the underlying distribution of $\epsilon_{\text{test}}(\hat{b})$ suggests their use.

These curves (Fig.~\ref{fig:optimized_sampling}) provide a comprehensive visualization of VIBES strategies performances across various time budgets, enabling nuanced comparisons. More precisely, BSEC can provide the following key insights:
\begin{itemize}[leftmargin=*, itemsep=5pt, parsep=0pt]
    \item The curve's shape indicates how quickly the backbones improve over time. A steep initial slope suggests rapid improvement with small time budgets.
    \item When curves are displayed until convergence, their plateau reveals the upper bound performance achievable with exhaustive search.
    \item The width of the area between percentiles represents the strategy's reliability across multiple runs. Narrower bands indicate more consistent performance.
    \item Comparing curves at specific time $t$ reveals which strategies are more effective for different time budgets. 
\end{itemize}

\vspace{-13pt}
\section{Proposed VIBES strategies}\label{sec:approach}
\vspace{-5pt}
To illustrate the concepts presented, we propose and compare different strategies encompassing both families of solutions: fast approximate evaluation and optimized sampling. These strategies serve as strong initial VIBES baselines for low-data image classification.

\vspace{-6pt}
\subsection{Fast approximate evaluation strategy}\label{sec:model_eval_approx}
For fast approximate evaluation, we propose to 
work directly in the feature space, by replacing learning-based classifiers with learning-free models. This approach reduces evaluation time by operating directly on extracted features and eliminating the classifier training phase.

In our implementation, we replace logistic regression by a nearest centroid algorithm. This learning-free classifier assigns to each validation sample the label of the training class whose centroid (mean feature vector) is closest to the sample's feature vector. 
Nearest centroid classifiers are particularly well-suited for low-data regimes as they avoid overfitting through their simplicity and parameter-efficiency. They also have low computational complexity for small number of samples: $\mathcal{O}(nd)$ during training and $\mathcal{O}(Cd)$ during inference, where $n$ is the number of samples, $d$ the feature dimension, and $C$ the number of classes. 

In Appendix A, we show experimentally that nearest centroid is significantly faster to compute than logistic regression. We also demonstrate that while nearest centroid performance is typically lower than logistic regression, the two metrics are strongly correlated across backbones. Hence, it appears to be a good choice to speed up backbone selection.

Finally, we also explore K-nearest neighbors (K=5) as an alternative learning-free classifier. Results are presented in Appendix B. 

\vspace{-6pt}
\subsection{Optimized sampling strategies}\label{sec:model_sampling_approx}
Optimized sampling aims to prioritize potentially high-performing backbones early in the search process. We propose five strategies:
\begin{enumerate}[leftmargin=*, itemsep=5pt, parsep=0pt]
    \item \textbf{Random} (baseline): uses a random permutation of backbones, 
    
    \item \textbf{Increasing model complexity}: tests small backbones first to evaluate more models within time constraints but may miss high-performing large models, 
    
    \item \textbf{Decreasing model complexity}: prioritizes large models that often perform better but evaluates fewer models overall,
    
    \item \textbf{Decreasing dataset size}: tests backbones pretrained on the largest datasets first, assuming more pretraining data yields better features but may overlook models trained on smaller yet more relevant datasets, and 
    
    \item \textbf{Dataset cycling}: alternates between pretraining datasets to maximize representation diversity but may not fully exploit the benefits of a single highly relevant dataset.
\end{enumerate}

These strategies represent different hypotheses about what factors most influence backbone performance. By comparing them, we aim to gain insights into the relative importance of model size, pretraining dataset size, and diversity for low-data image classification.

\section{Experiments}
\label{sec:experiments}

Our experiments focus on image classification with limited data to showcase the potential of dataset-specific backbone selection. While VIBES is applicable to various computer vision problems, low-data classification provides a clear demonstration of its effectiveness and implementation process. To ensure full reproducibility of our results, we provide a comprehensive GitHub repository\footnote{\url{https://github.com/jorisguerin/vibes-fsl}} containing all instructions, code and configuration files.


\subsection{Datasets}
We conducted experiments on four datasets, representing different domains: 
\begin{itemize}[leftmargin=*, itemsep=5pt, parsep=0pt]
    \item \textbf{CIFAR10}~\cite{cifar10}, a widely used benchmark composed of $32\!{\times}\!32$ color images across 10 classes; 
    
    \item \textbf{GTSRB}~\cite{gtsrb} containing traffic sign images across 43 classes with sizes from $15\!{\times}\!15$ to $250\!{\times}\!250$ pixels and varying lighting conditions; 
    
    \item \textbf{Flowers102}~\cite{flowers102} consisting of 8,189 flower images across 102 categories with only 10 images per category for training, representing fine-grained classification with high inter-class similarity; and 
    
    \item \textbf{EuroSAT}~\cite{eurosat} containing remote sensing satellite images ($64\!{\times}\!64$) across 10 land cover classes, representing a domain shift from traditional pretraining datasets.
\end{itemize}

To simulate realistic low-data scenarios, we randomly sample $N=10$ examples per class from each training set. Additional experiments with varying $N$ are presented in Appendix~C. For evaluation, we use the full test splits for statistical significance. For EuroSAT, since no official split is available, we create custom splits. Split indices are included in the source code to ensure full reproducibility.

\subsection{Implementation details}
We apply VIBES strategies to a large set of $1,322$ vision backbones from timm~\cite{rw2019timm}, representing a wide diversity of architectures (ConvNets, Vision Transformers, MLP-Mixers, etc.) and sizes (from $\sim$500k parameters to over 1.2 billion). The complete list of backbones can be found in our GitHub repository.

For each backbone evaluation, we first preprocess the input images according to the specific backbone requirements. Then, we extract the pre-classifier features and train either a logistic regression classifier or a nearest centroid classifier depending on the evaluation strategy. For logistic regression, we use the LBFGS optimizer with L2 regularization (C=1.0) and a maximum of 100 iterations. For Nearest Centroid, we use Euclidean distance. All experiments are conducted on a single NVIDIA RTX A5000 GPU with 24GB memory.

Our backbone selection approach maintains a simple evaluation procedure to enable rapid assessment of many backbones. While more elaborate training (e.g., using data augmentation) could improve asymptotic performance, it would significantly increase the computational cost of running VIBES. More advanced fine-tuning techniques can be applied to the selected model afterwards.

\vspace{-10pt}
\section{Results}
\vspace{-5pt}

We conducted extensive experiments to evaluate different VIBES heuristics, including both fast approximate evaluation and optimized sampling strategies. The results for fast approximate evaluation, which consists of replacing logistic regression with nearest centroid or k-nearest neighbors, are presented in Appendix B, along with detailed analysis. Unfortunately, this family of approaches was rather inconclusive and did not provide significant advantages over parametric classifiers.

\begin{figure}[t]
    \centering
    \begin{minipage}[b]{0.48\textwidth}
        \includegraphics[width=\textwidth]{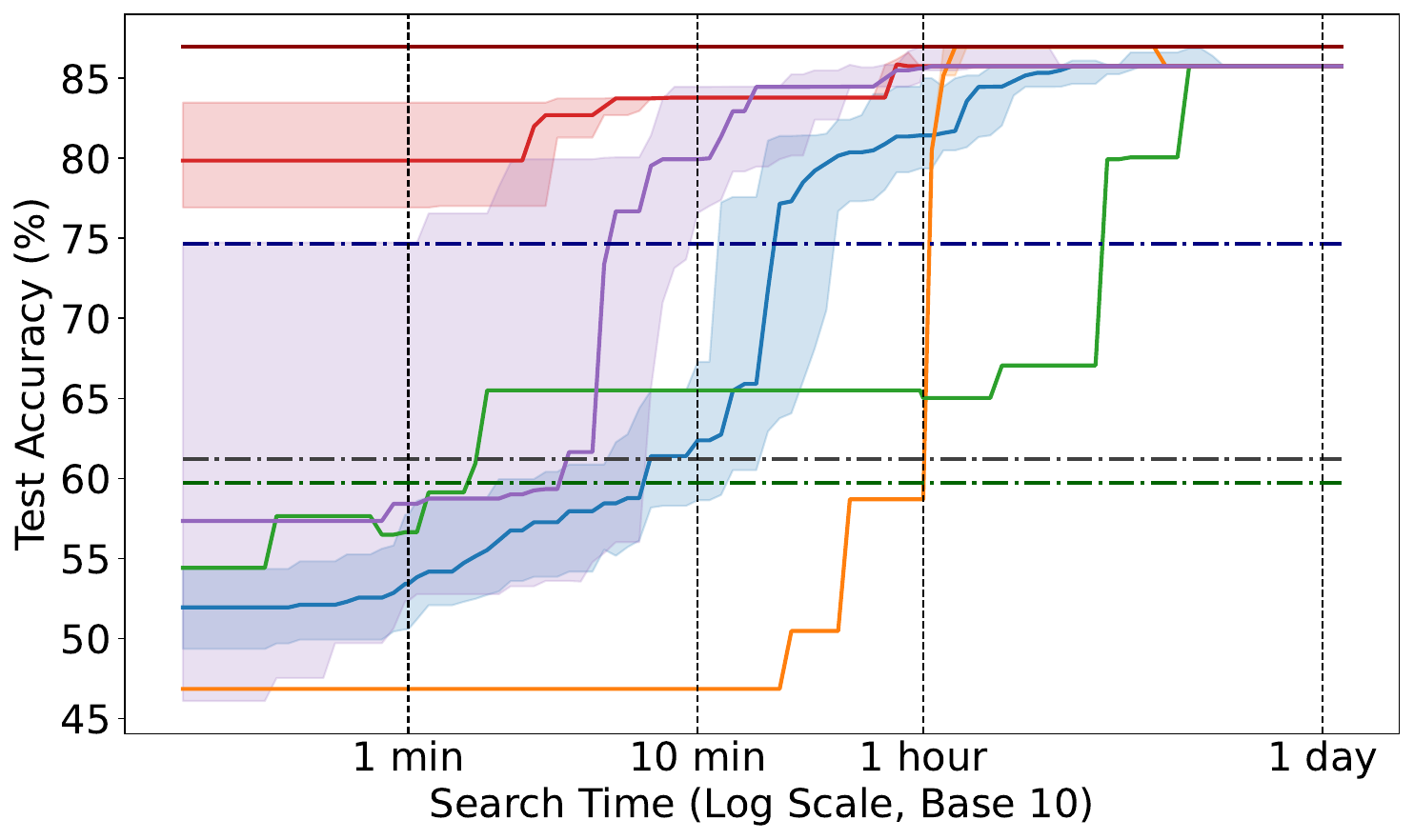}
        \parbox[t]{\textwidth}{\centering\small (a) GTSRB}
    \end{minipage}%
    \hfill
    \begin{minipage}[b]{0.48\textwidth}
        \includegraphics[width=\textwidth]{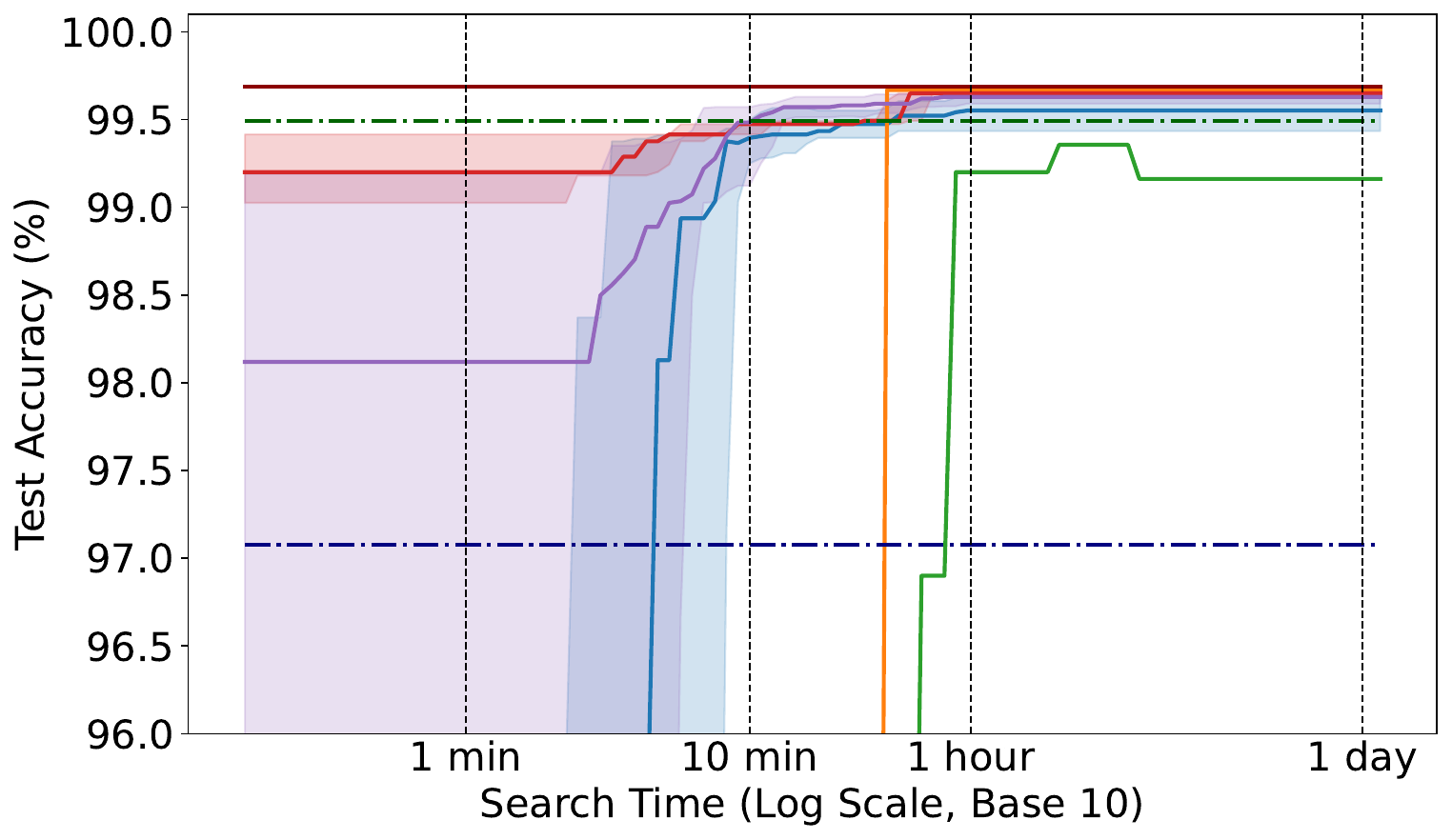}
        \parbox[t]{\textwidth}{\centering\small (b) Flowers102}
    \end{minipage}
    
    \vspace{0.1cm}
    
    \begin{minipage}[b]{0.48\textwidth}
        \includegraphics[width=\textwidth]{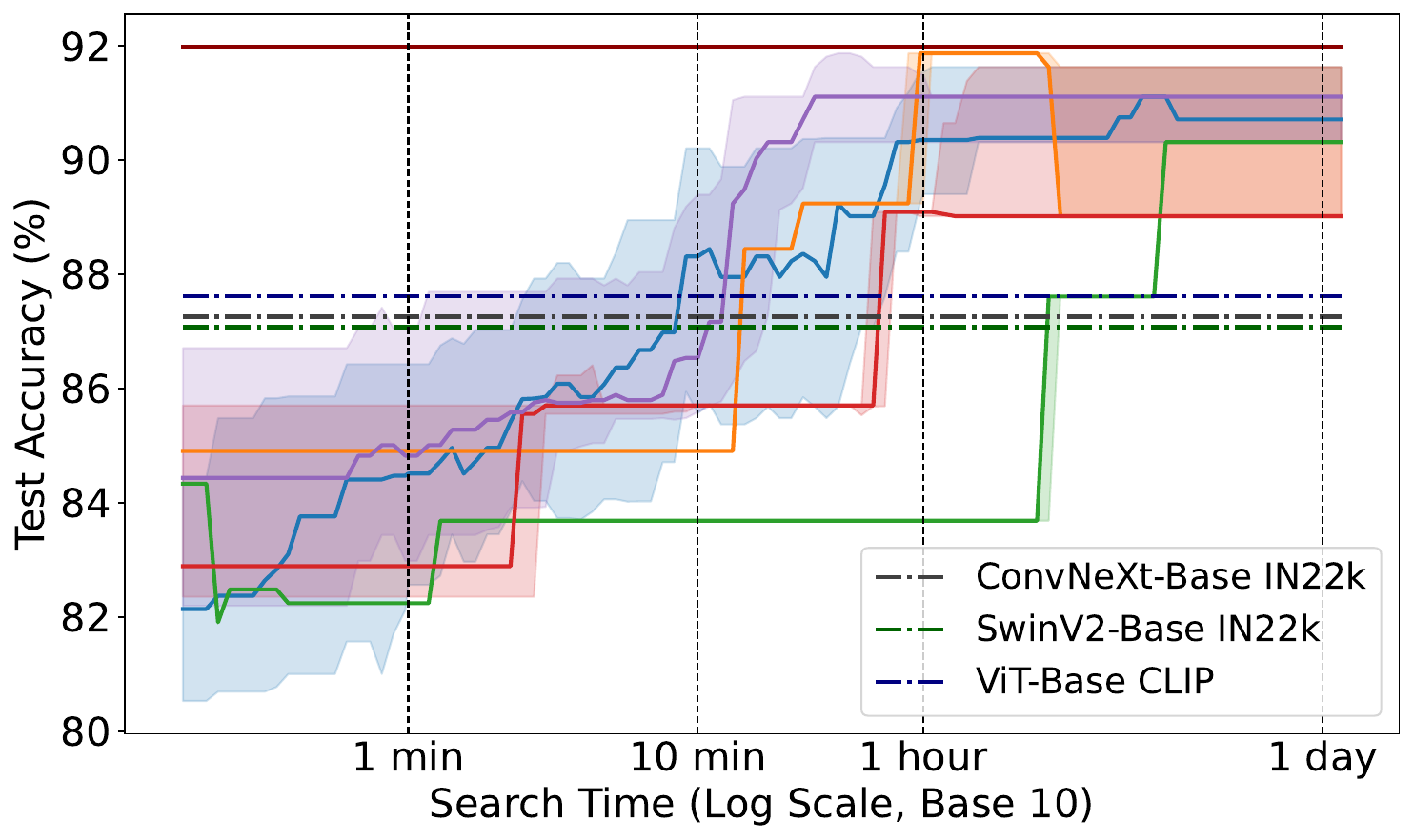}
        \parbox[t]{\textwidth}{\centering\small (c) EuroSAT}
    \end{minipage}%
    \hfill
    \begin{minipage}[b]{0.48\textwidth}
        \includegraphics[width=\textwidth]{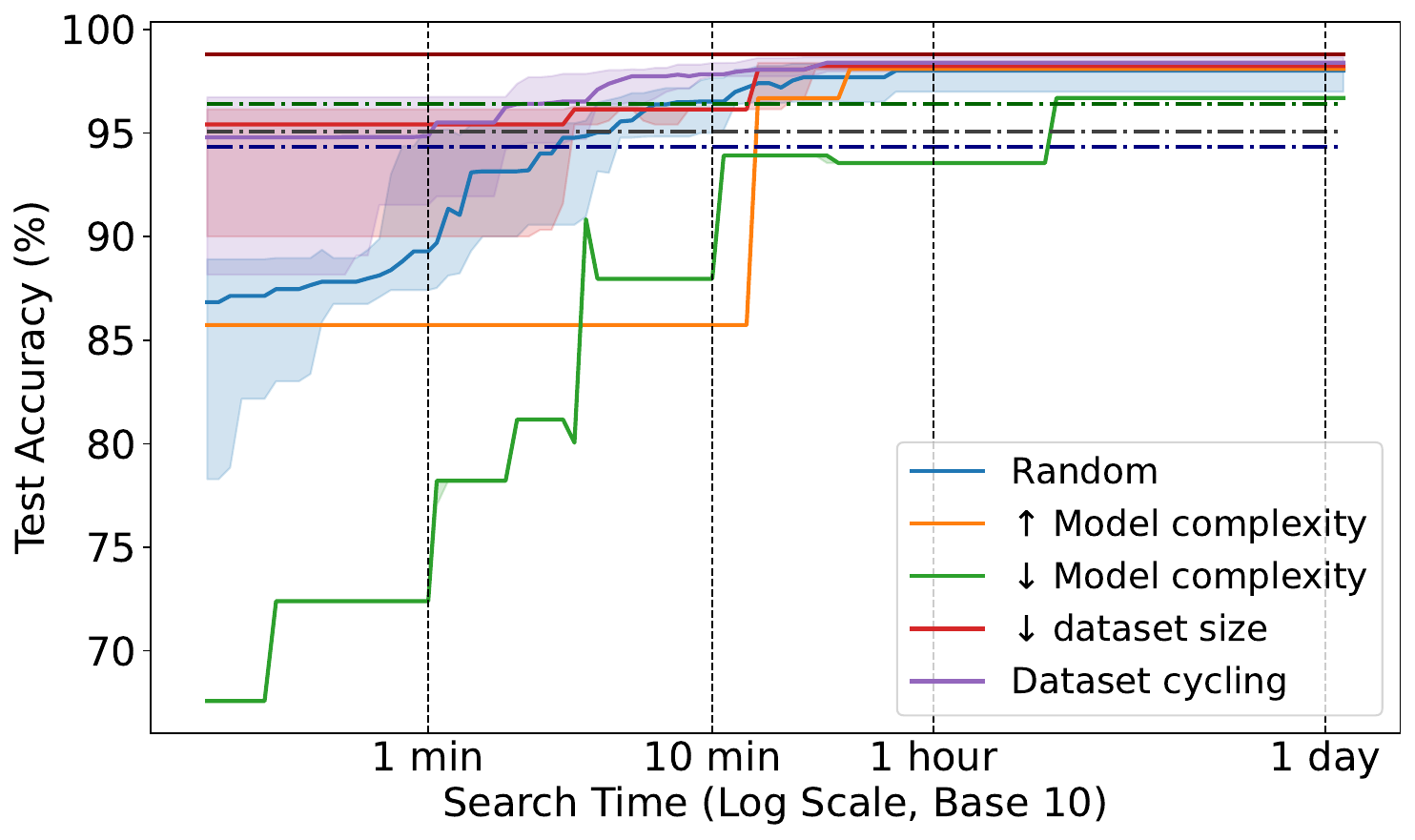}
        \parbox[t]{\textwidth}{\centering\small (d) CIFAR10}
    \end{minipage}
    
    \caption{\textbf{Optimized sampling strategies}. Backbone Selection Efficiency Curves comparing different methods for defining backbone evaluation order: random sampling (baseline), ordering by increasing/decreasing model complexity, ordering by decreasing pretraining dataset size, and dataset cycling. All curves use logistic regression as the classifier.}
    \label{fig:optimized_sampling}
\end{figure}

Therefore, in the main paper, we focus on comparing the performance of VIBES strategies based on optimized sampling. Figure~\ref{fig:optimized_sampling} presents these results across four datasets. In all our results figures, as a baseline, we also display the performance of the three backbones recommended by the most recent benchmark study~\cite{goldblum2024battle}. All experiments use logistic regression during the backbone evaluation phase in a low-data context with N=10 samples per class. The complete results for all backbones tested under different configurations are available in CSV format in our GitHub repository.

\subsection{dataset-specific selection vs. benchmark results} 

The first objective of our experiments is to determine whether VIBES can identify dataset-specific backbones that outperform recommended general-purpose backbones in low-data regimes. Comparing the general-purpose backbone baselines against the blue curves (representing simple random sampling with logistic regression evaluation), we observe:
\begin{enumerate*}
\item Within just ten minutes of search time, the most basic VIBES strategy outperforms nearly all general-purpose backbone models across datasets.
\item With one hour of search, this basic VIBES approach significantly surpasses benchmark-recommended backbones: outperforming ConvNeXt/SwinV2/ViT by approximately 20\%/22\%/7\% for GTSRB, 0.1\%/0.1\%/2.5\% for Flowers102 , 3\%/3\%/2.5\% for EuroSAT, and 3.5\%/2\%/4.5\% for CIFAR10.
\end{enumerate*} 
These results highlight the clear advantage of dataset-specific backbone selection over one-size-fits-all approaches. This advantage is particularly pronounced in low-data settings, where the quality of the initially selected backbone is paramount since extensive fine-tuning is not feasible with limited training examples.

\subsection{Optimized sampling strategies comparison}
Figure~\ref{fig:optimized_sampling} presents results for our proposed optimized sampling strategies, with random sampling as a baseline for comparison.

Sampling strategies based on model complexity, whether increasing or decreasing, consistently underperform compared to random sampling across nearly all datasets and time budgets. This outcome suggests that a backbone's potential to perform well on target datasets does not correlate with its number of trainable parameters, challenging the intuitive assumption that large models offer better transferability. We note that both decreasing and increasing model complexity strategies show high reliability across multiple runs due to their deterministic nature, as evidenced by their narrow confidence bands.

In contrast, sampling methods based on pre-training datasets demonstrate more promising results. Both dataset cycling and decreasing dataset size strategies outperform random sampling for GTSRB, CIFAR10, and Flowers102. The effectiveness of these approaches underscores the importance of considering the nature and diversity of pre-training data when selecting backbones for low-data transfer learning. 

Interestingly, for EuroSAT (satellite imagery), decreasing dataset size performs poorly, while dataset cycling maintains performance comparable to random sampling. This discrepancy highlights that pre-training dataset size alone is not a reliable predictor of transferability. It suggests a more nuanced relationship where the relevance of the pre-training dataset to the target domain is equally, if not more, important than its size. This observation might be particularly relevant for domains like remote sensing that differ substantially from natural images commonly used in pre-training.

To evaluate the robustness of these findings to different numbers of training examples (N), Appendix C provides a detailed experimental comparison between random sampling and dataset cycling with varying values of N.



\section{Conclusion}
\label{sec:conclusion}
 
The main contribution of this work is to formalize Vision Backbone Efficient Selection (VIBES). We introduce several strategies to solve this problem and apply them to image classification tasks under limited available data. Low-data regimes are particularly well-suited for backbone selection optimization, as initial feature quality becomes paramount when insufficient data is available to fine-tune representations.

Our experiments yield compelling results, showing that simple VIBES approaches can identify backbones that outperform general-purpose benchmarks within just ten minutes of search time on a single GPU. Among the strategies tested, faster approximate evaluation methods (replacing logistic regression with nearest centroid) did not deliver the expected benefits. However, optimizing backbone sampling by prioritizing diversity in pretraining datasets through our dataset cycling approach, showed consistent improvements for backbone selection across different datasets, time budgets and levels of data scarcity.


These findings open several promising avenues for future research:
\begin{enumerate}[leftmargin=*, itemsep=5pt, parsep=0pt]
    \item Developing new VIBES strategies that combine approximate evaluation with optimized sampling or leverage meta-learning techniques to predict backbone performance without full evaluation.
    \item Extending experiments to other vision tasks such as object detection, semantic segmentation, and pose estimation to better delineate scenarios where backbone search offers significant advantages.
    \item Building a user-friendly platform where practitioners could upload their small datasets and receive optimized backbone recommendations. This would not only facilitate adoption but also enable collection of valuable data on backbone transferability across diverse domains, informing the development of more sophisticated sampling strategies grounded in empirical performance patterns.
\end{enumerate}


{\small
\bibliography{main}
}

\newpage
\appendix
\section*{Appendix}
In this appendix, we present additional experiments to evaluate other aspects of vision backbone efficient selection (VIBES) for low-data image classification.

\section{Comparing nearest centroid performance against logistic regression}

A key component of our fast approximate evaluation strategy is the use of nearest centroid classification as a proxy for logistic regression. Here, we analyze the relationship between these methods in terms of both computational efficiency and predictive performance.

\subsection{Computational efficiency}
Nearest centroid demonstrates significant computational advantages over logistic regression in our experimental setup. In a 10 samples per class setting, averaging across all 1,300+ backbones, we get the following training and evaluation times:
Logistic regression: 220 milliseconds; Nearest centroid: 3 milliseconds. This important speedup can be beneficial to the backbone selection process where multiple evaluations are needed.

\subsection{Performance correlation}
\begin{figure}[!ht]
    \centering
    \includegraphics[width=0.9\linewidth]{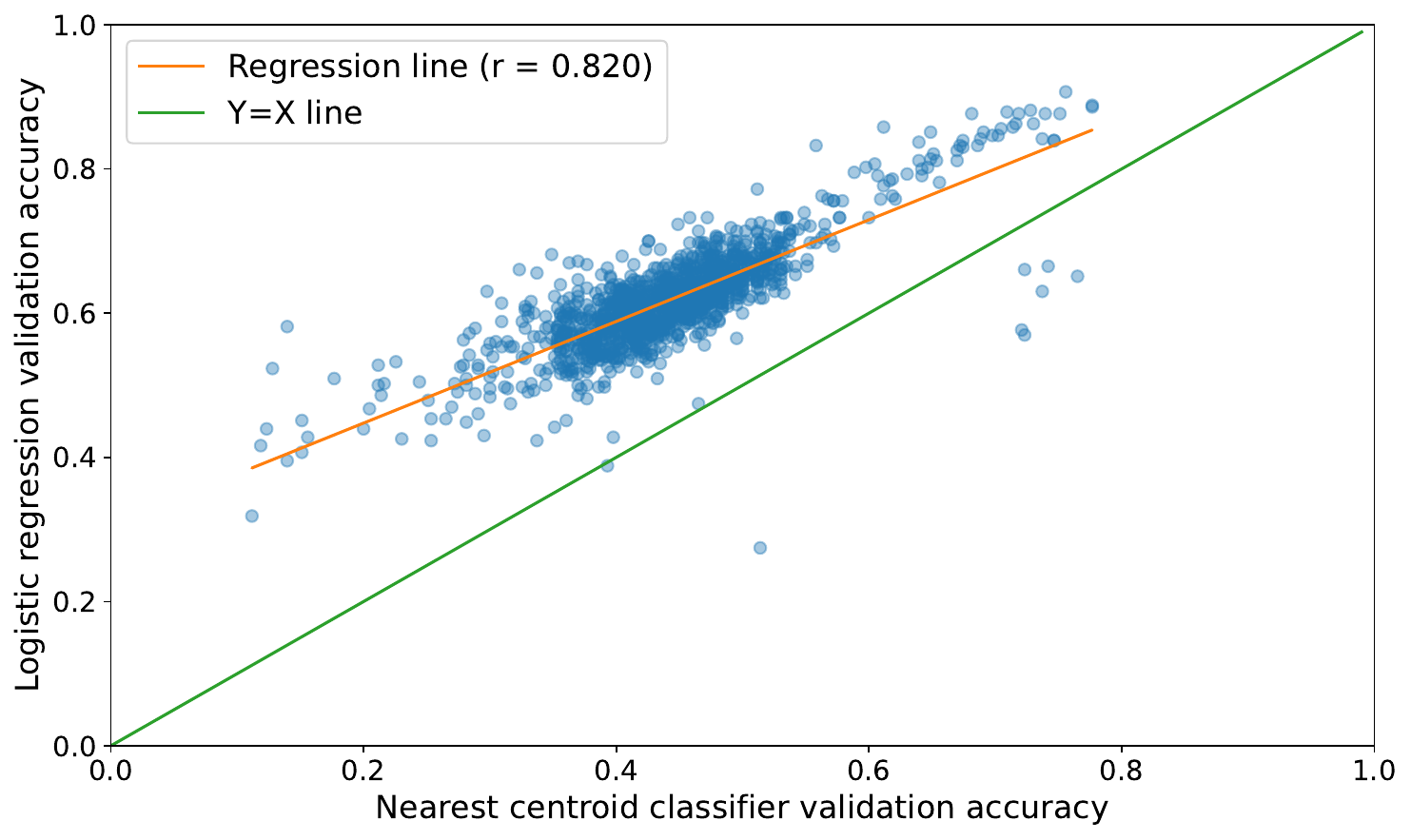}
    \caption{{Nearest centroid vs. logistic regression performance on low-data GTSRB.} Each point represents a pretrained backbone evaluated using both classifiers with 10 samples per class. Points above the identity line indicate logistic regression outperforming nearest centroid. The strong linear correlation demonstrates that nearest centroid effectively ranks backbones despite lower absolute performance.}
    \label{fig:NCvsLOG_gtsrb}
\end{figure}

\begin{figure}[!ht]
    \centering
    \includegraphics[width=0.9\linewidth]{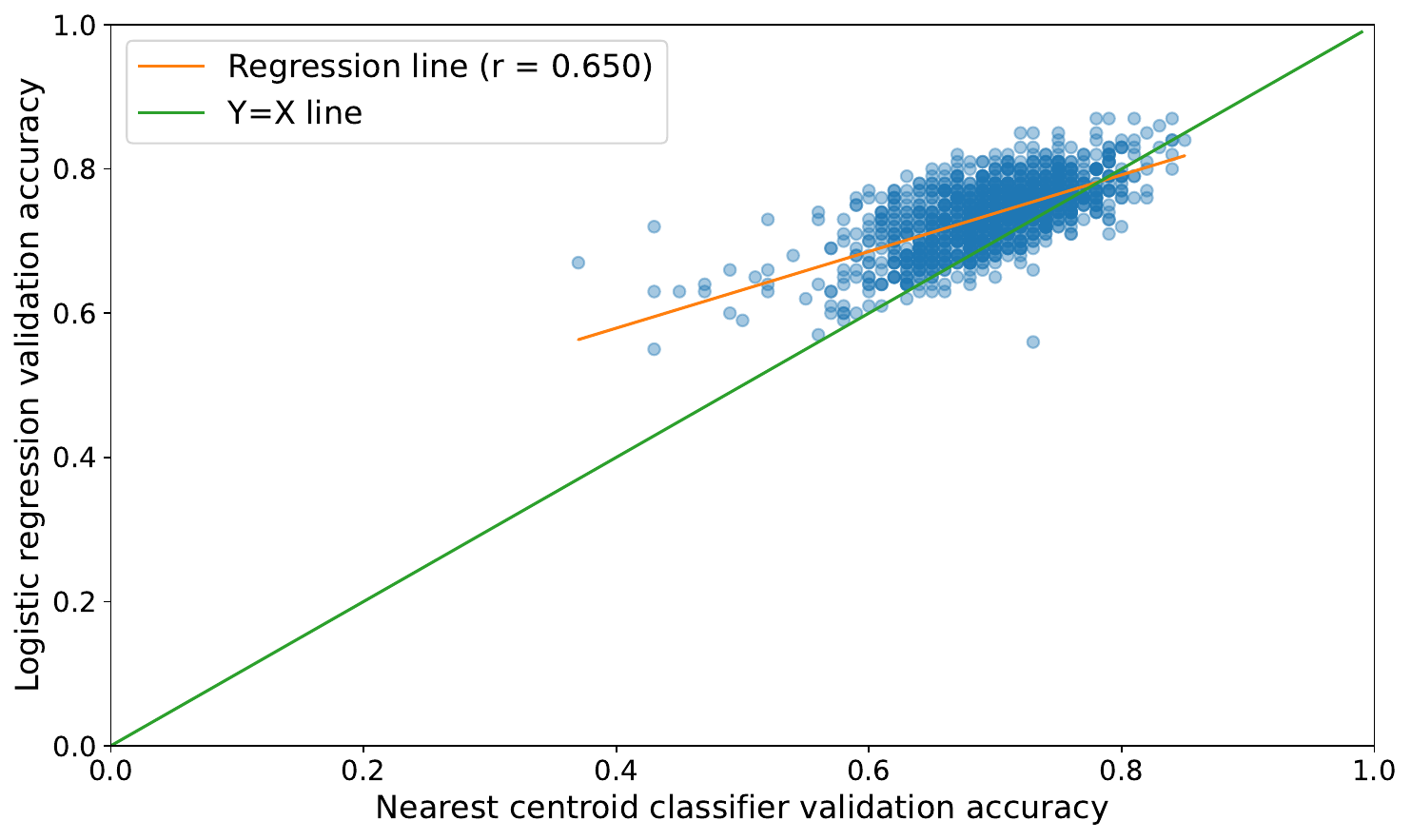}
    \caption{{Nearest centroid vs. logistic regression performance on low-data EuroSAT.}}
    \label{fig:NCvsLOG_eurosat}
\end{figure}

\begin{figure}[!ht]
    \centering
    \includegraphics[width=0.9\linewidth]{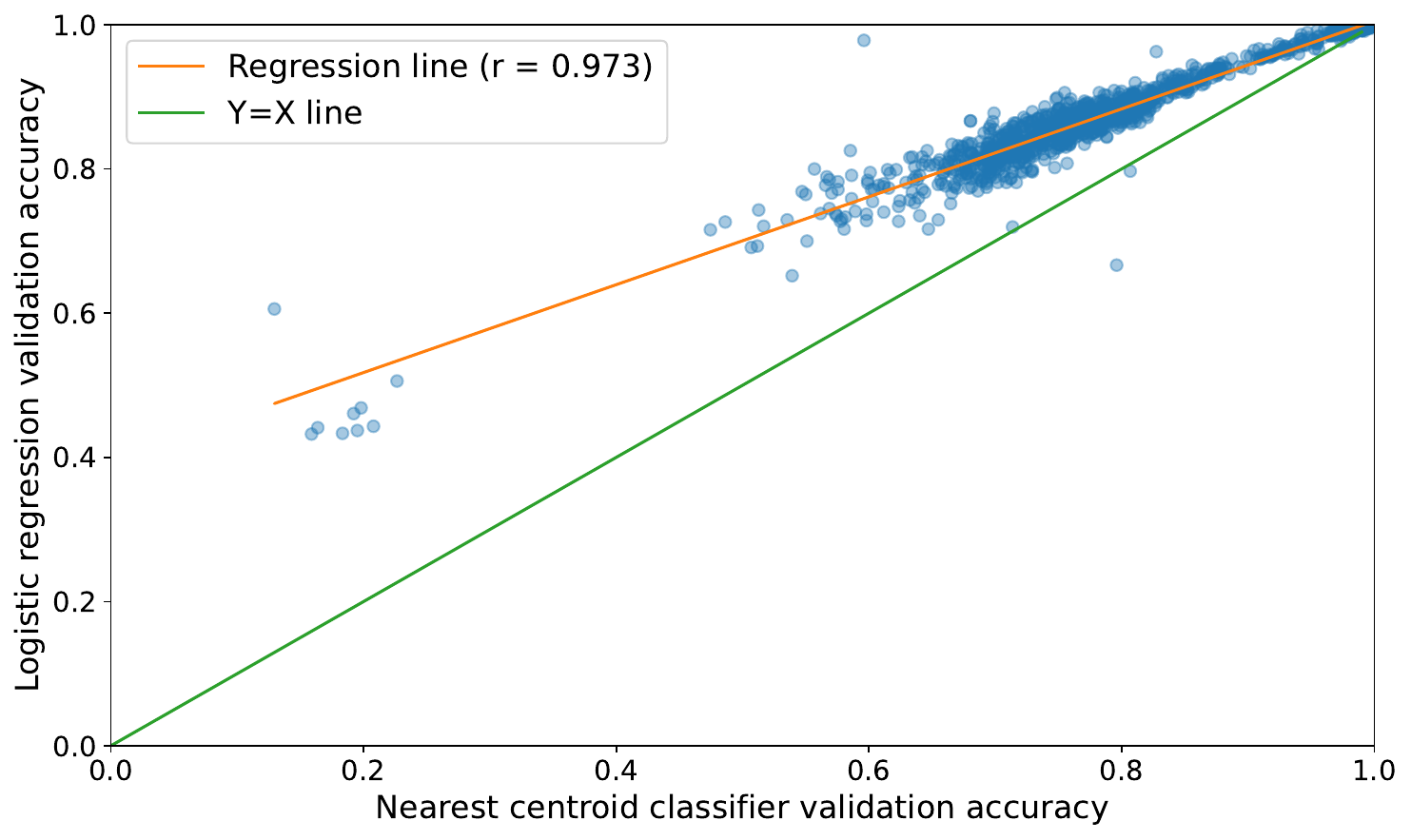}
    \caption{{Nearest centroid vs. logistic regression performance on low-data Flowers102.}}
    \label{fig:NCvsLOG_flowers102}
\end{figure}

\begin{figure}[!ht]
    \centering
    \includegraphics[width=0.9\linewidth]{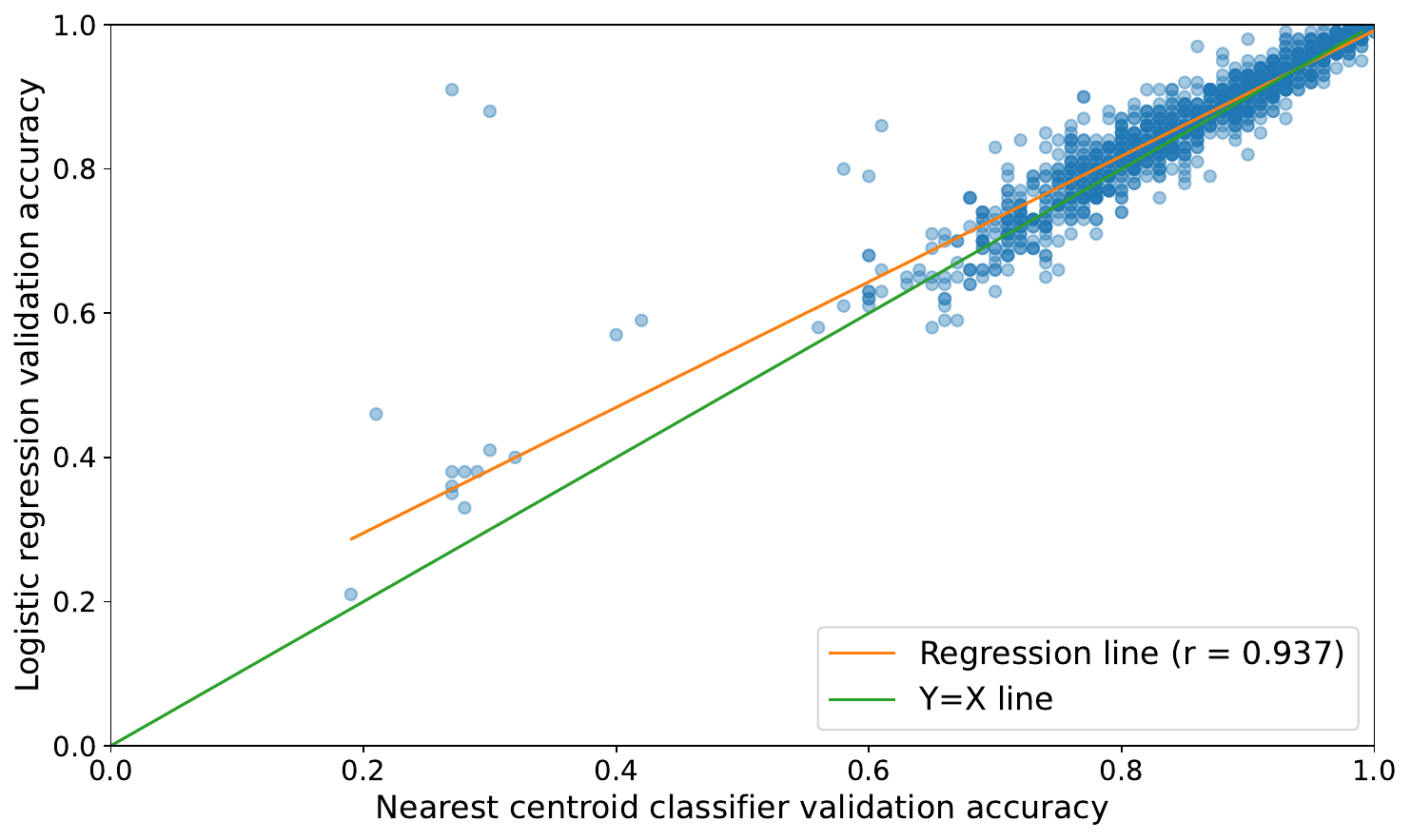}
    \caption{{Nearest centroid vs. logistic regression performance on low-data CIFAR10.}}
    \label{fig:NCvsLOG_cifar10}
\end{figure}

Figures~\ref{fig:NCvsLOG_gtsrb}–\ref{fig:NCvsLOG_cifar10} compare the validation accuracy of nearest centroid versus logistic regression across all backbones for each dataset.
Two key observations emerge:
\begin{itemize}[leftmargin=*, itemsep=5pt, parsep=0pt]
    \item \textbf{Performance gap}: Logistic regression consistently outperforms nearest centroid (points above the identity line), justifying our choice of using it as the final classifier.

    \item \textbf{Strong correlation}: The performance metrics show high correlation across backbones, indicating that nearest centroid serves as an effective proxy for ranking backbones despite its lower absolute performance.
\end{itemize}
These findings support our approach of using nearest centroid during the search phase to rapidly identify promising backbones, then applying logistic regression to the selected backbone for final model training.

\FloatBarrier
\newpage
\section{Assessing and comparing fast approximate evaluation strategies}
\begin{figure}[!ht]
    \centering
    \begin{minipage}[b]{0.48\textwidth}
        \includegraphics[width=\textwidth]{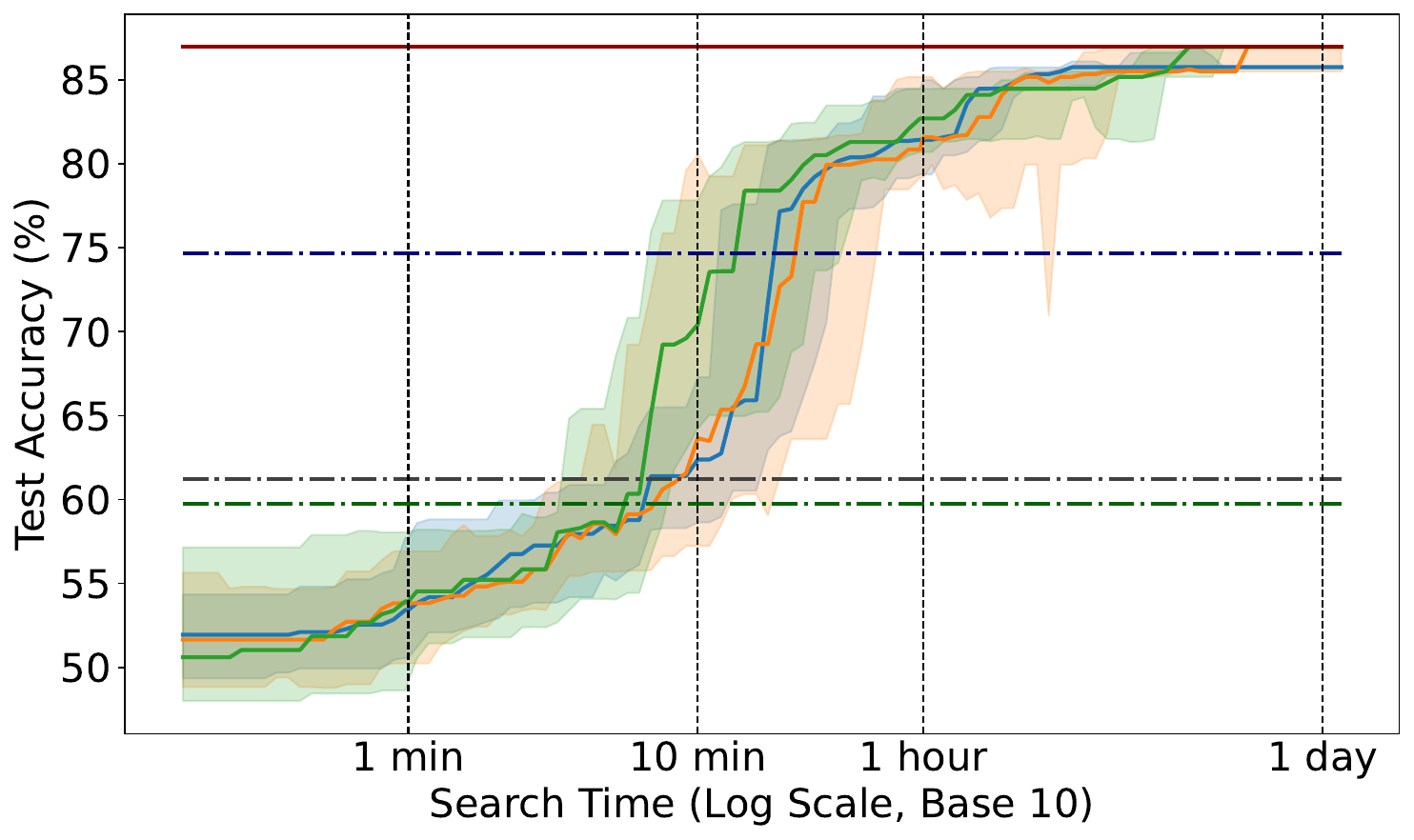}
        \parbox[t]{\textwidth}{\centering\small (a) GTSRB}
    \end{minipage}%
    \hfill
    \begin{minipage}[b]{0.48\textwidth}
        \includegraphics[width=\textwidth]{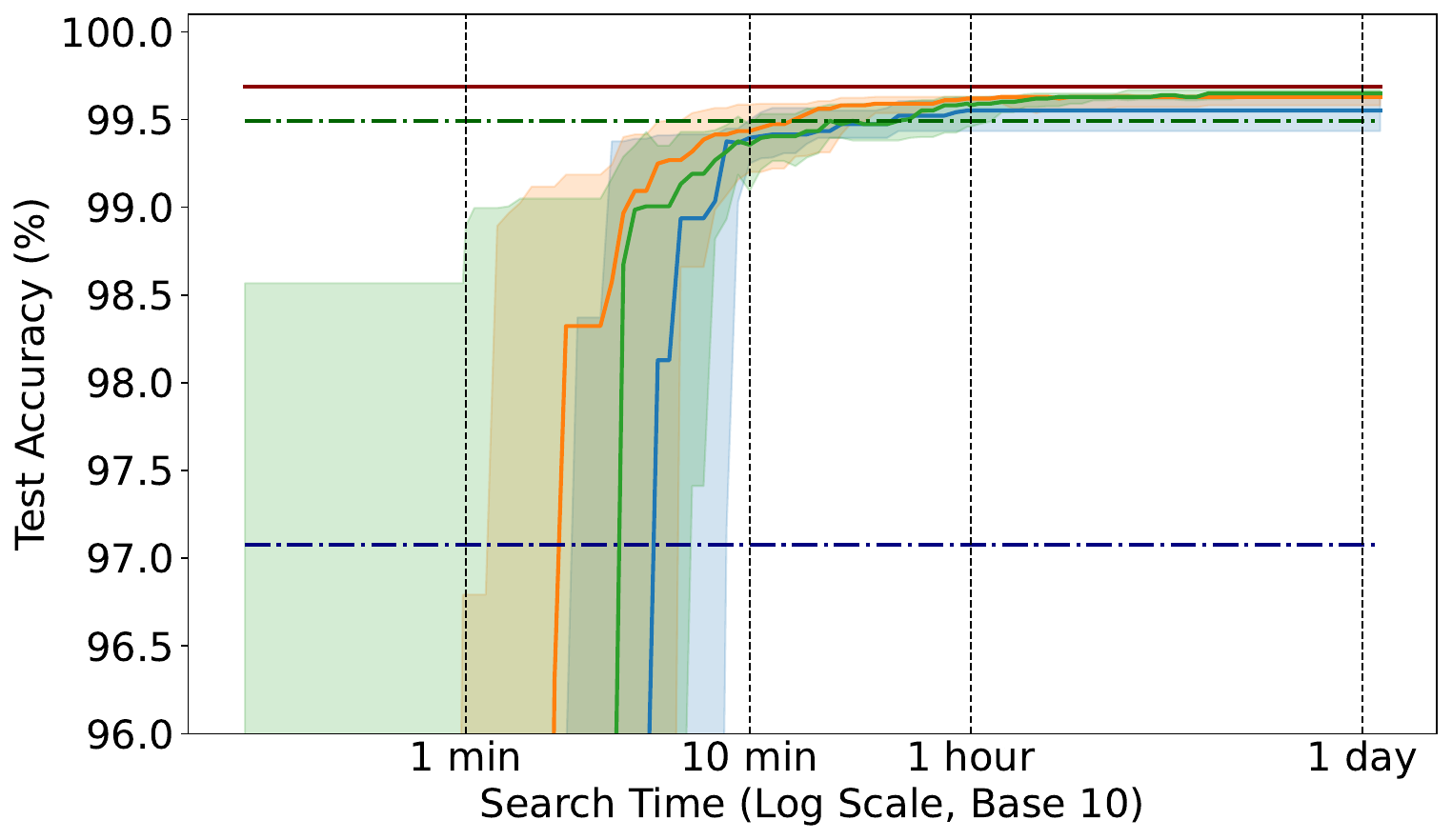}
        \parbox[t]{\textwidth}{\centering\small (b) Flowers102}
    \end{minipage}
    
    \vspace{0.1cm}
    
    \begin{minipage}[b]{0.48\textwidth}
        \includegraphics[width=\textwidth]{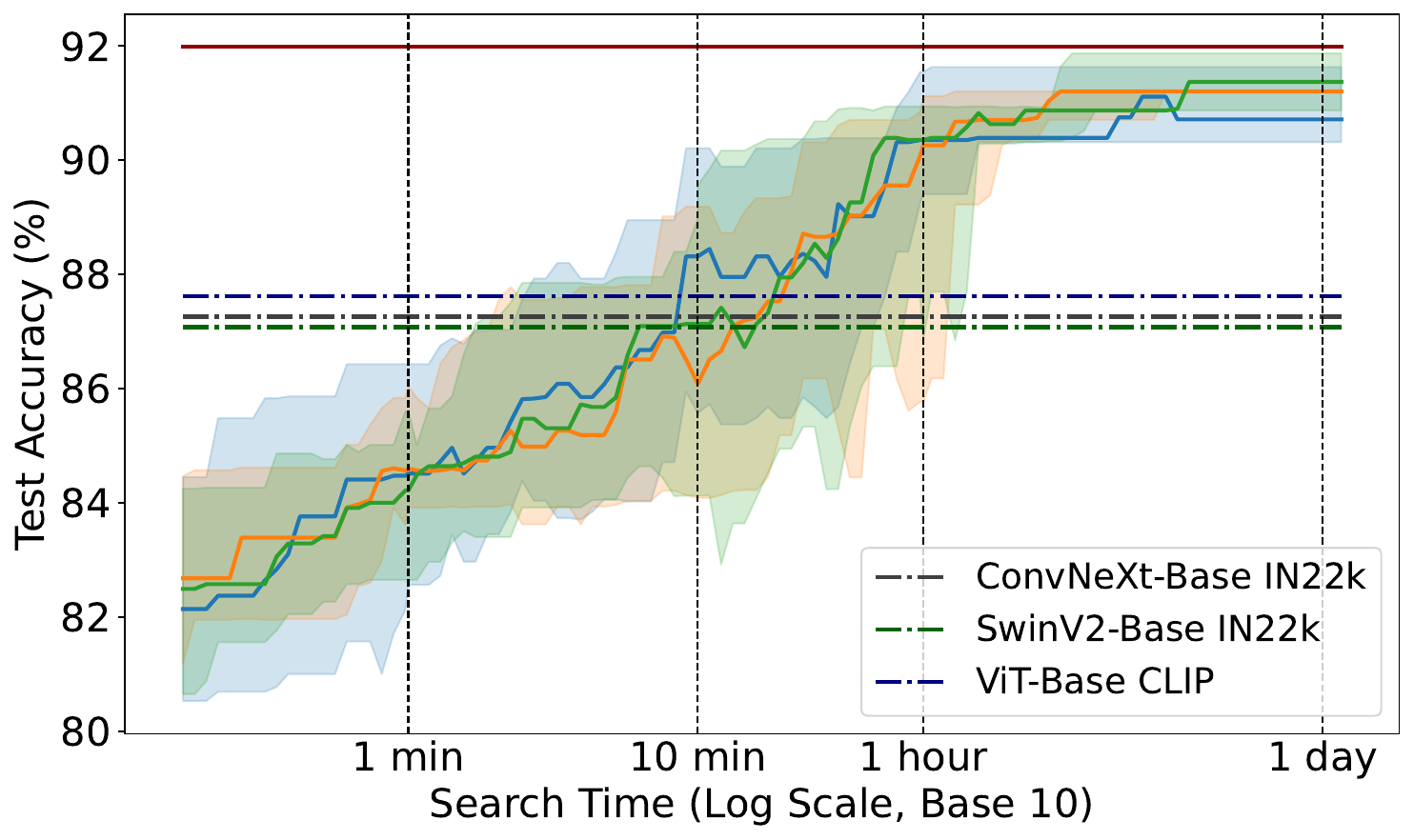}
        \parbox[t]{\textwidth}{\centering\small (c) EuroSAT}
    \end{minipage}%
    \hfill
    \begin{minipage}[b]{0.48\textwidth}
        \includegraphics[width=\textwidth]{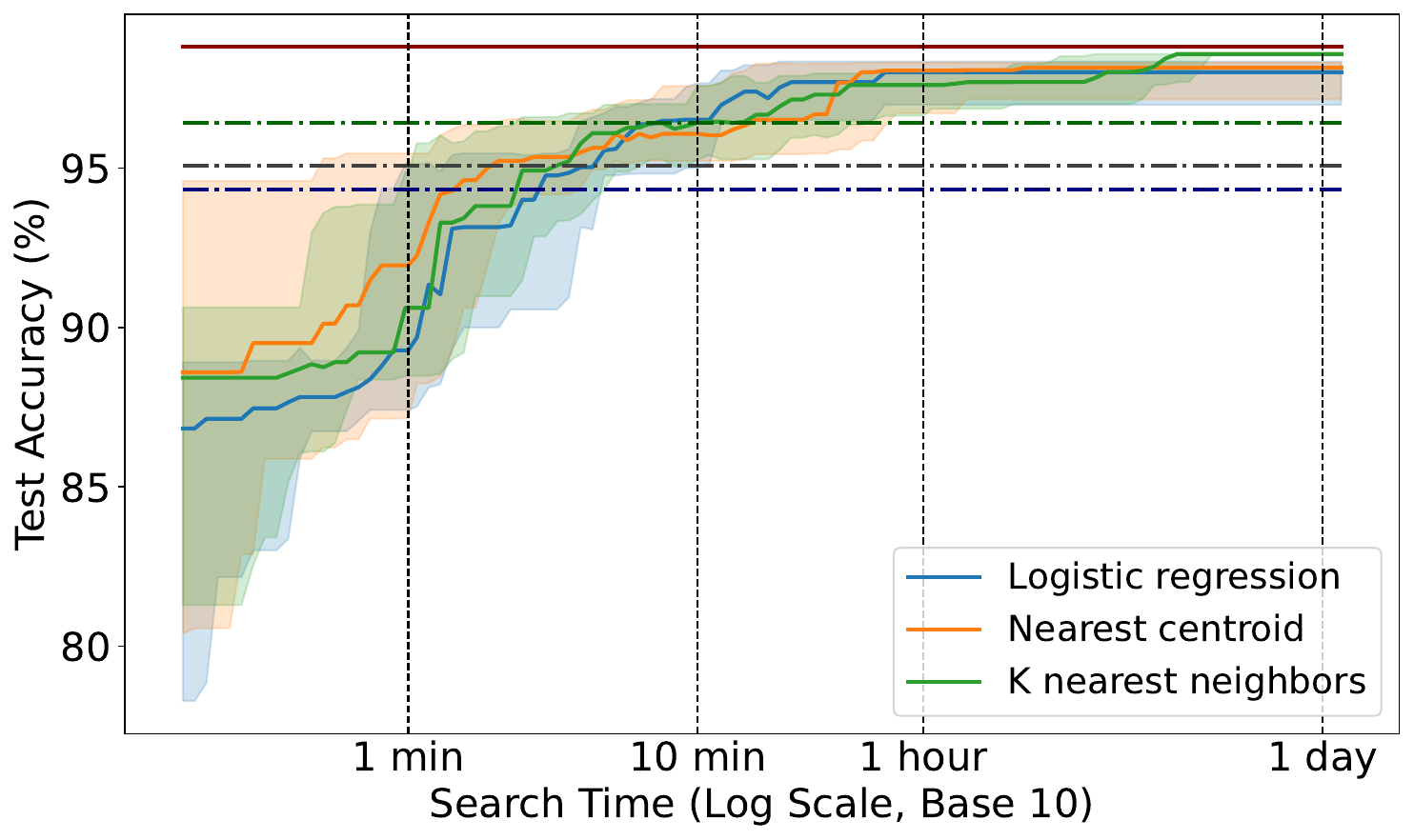}
        \parbox[t]{\textwidth}{\centering\small (d) CIFAR10}
    \end{minipage}
    
    \caption{\textbf{Fast approximate evaluation strategies}. Backbone Selection Efficiency Curves comparing logistic regression, nearest centroid, and k-nearest neighbors (k=5) for backbone evaluation. All curves are computed under random backbone sampling order.}
    \label{fig:fast_evaluation}
\end{figure}

As a potential direction for improved vision backbone selection, we explored replacing logistic regression with faster classification alternatives. Specifically, we compared logistic regression against nearest centroid and k-nearest neighbors (k=5). Figure~\ref{fig:fast_evaluation} displays Backbone Selection Efficiency Curves (BSEC) for these strategies across four datasets.

Our results indicate that the performance of all three approaches follow very similar pattern. The minor differences observed lack statistical significance. However, nearest centroid and k-nearest neighbors, exhibit slightly increased variability and occasionally yield suboptimal selections (see the lower interval curves for GTSRB, EuroSAT). 

This underperformance might be explained by an unfavorable trade-off between evaluation speed and ranking accuracy. While training is significantly faster with non-parametric models, these computational savings represent only a small fraction of the total evaluation time, which is dominated by backbone downloading and feature extraction. Consequently, the minor ranking errors introduced by approximate evaluation sometimes outweigh their speed advantages.

\FloatBarrier
\newpage
\section{Comparing optimized sampling strategies under different levels of data scarcity}
Our main analysis was conducted in a specific low-data regime with N=10 samples per class. To ensure that our findings generalize beyond this particular constraint, we conducted additional experiments on the GTSRB dataset across a range of data availability scenarios.

\begin{figure}[!ht]
    \centering
    \begin{minipage}[b]{0.48\textwidth}
        \includegraphics[width=\textwidth]{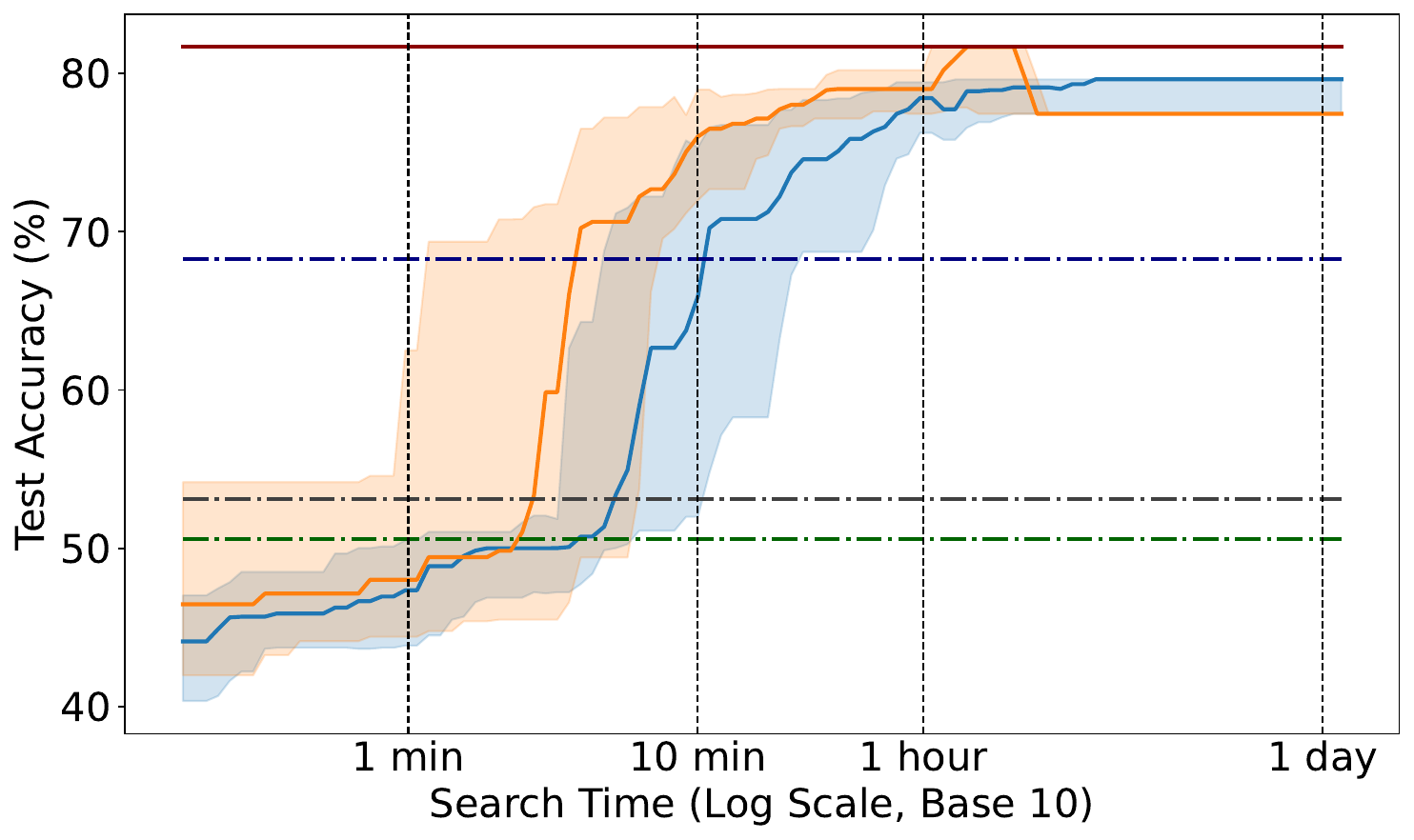}
        \parbox[t]{\textwidth}{\centering\small (a) $N=5$ sample per class}
    \end{minipage}%
    \hfill
    \begin{minipage}[b]{0.48\textwidth}
        \includegraphics[width=\textwidth]{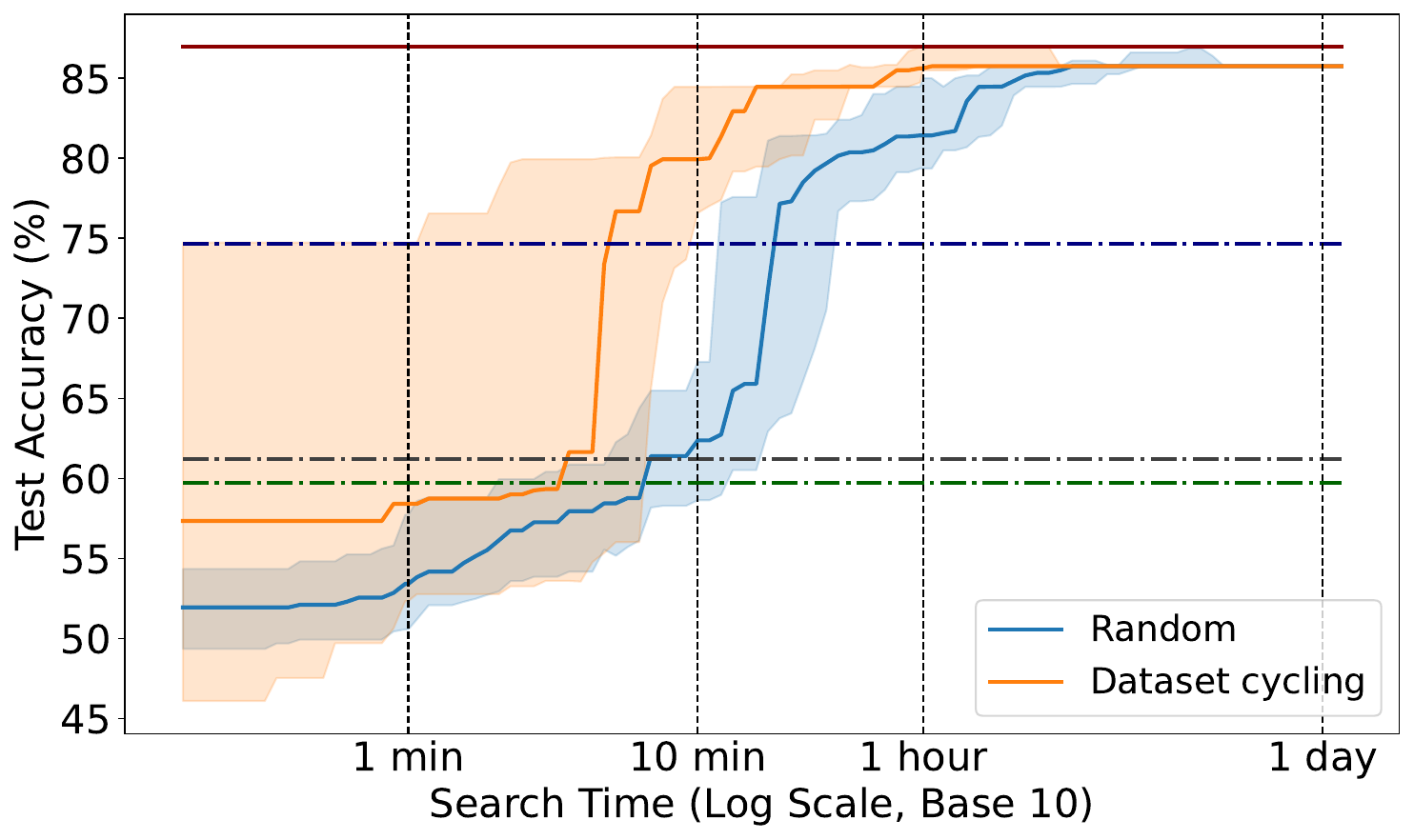}
        \parbox[t]{\textwidth}{\centering\small (b) $N=10$ sample per class}
    \end{minipage}
    
    \vspace{0.1cm}
    
    \begin{minipage}[b]{0.48\textwidth}
        \includegraphics[width=\textwidth]{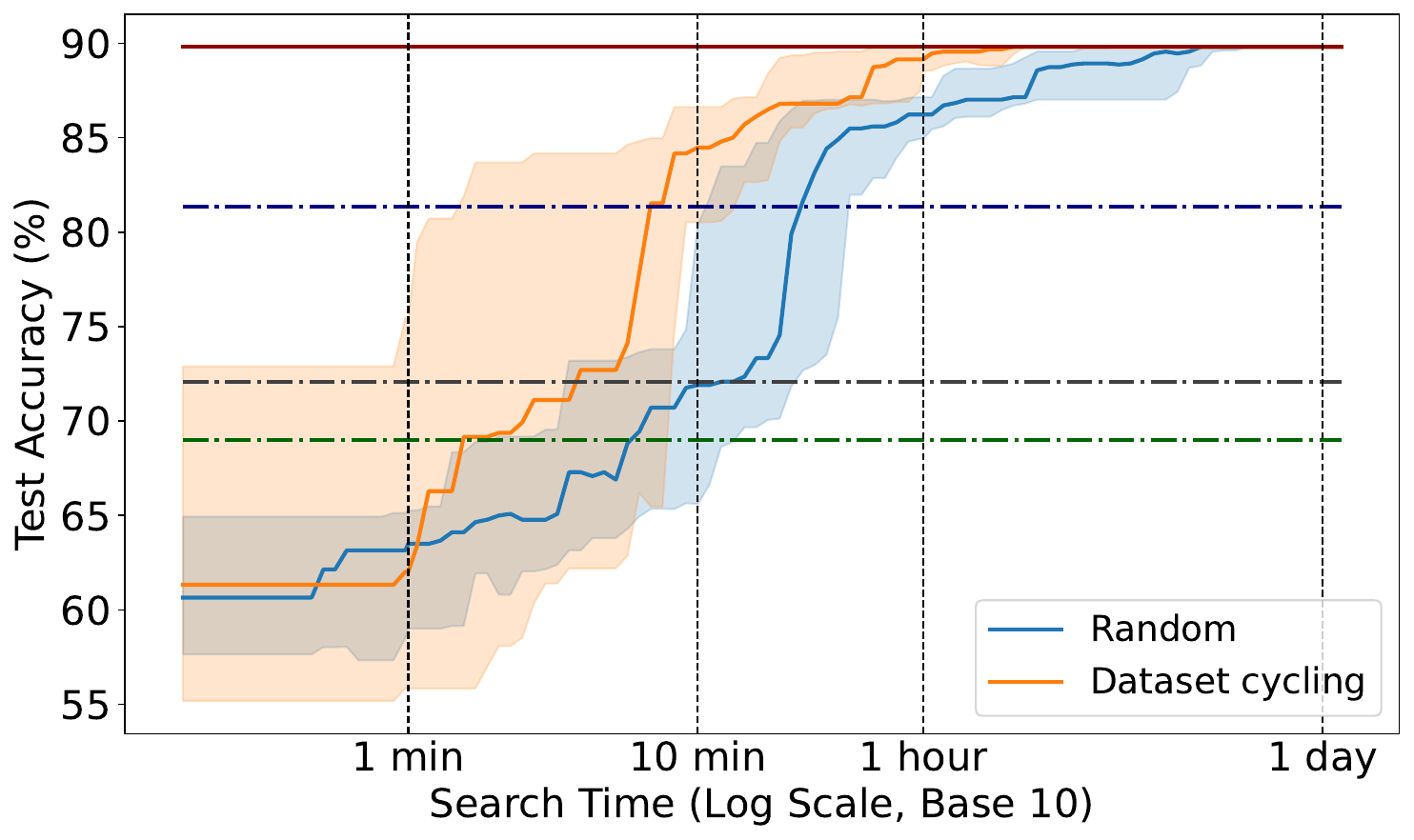}
        \parbox[t]{\textwidth}{\centering\small (c) $N=20$ sample per class}
    \end{minipage}%
    \hfill
    \begin{minipage}[b]{0.48\textwidth}
        \includegraphics[width=\textwidth]{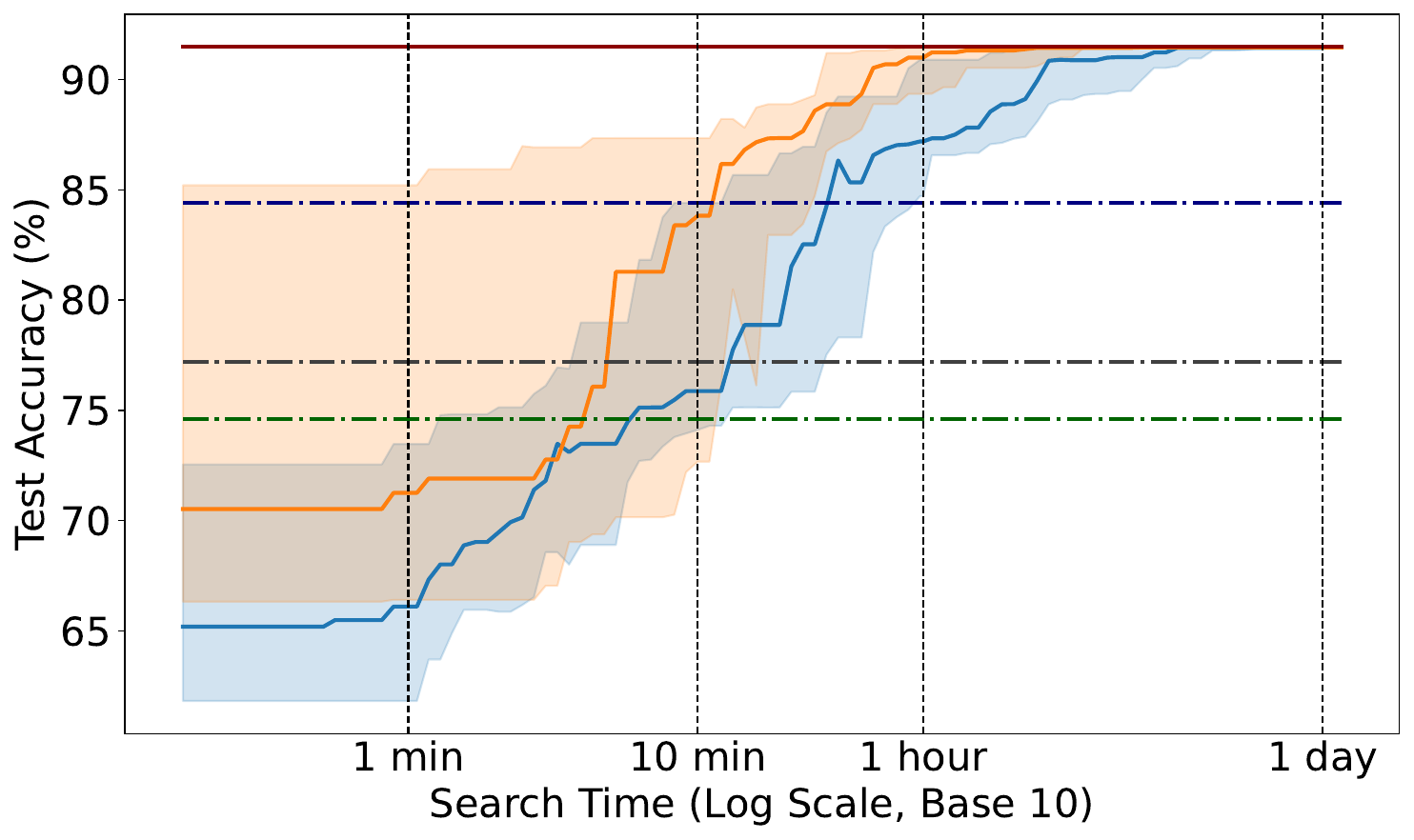}
        \parbox[t]{\textwidth}{\centering\small (d) $N=40$ sample per class}
    \end{minipage}
    
    \caption{\textbf{Comparing random sampling and dataset cycling under different levels of data scarcity (GTSRB dataset)}. Backbone Selection Efficiency Curves for random sampling and dataset cycling as the number of available training samples per class varies from very scarce (N=5) to less constrained (N=40). The consistent performance advantage of dataset cycling across these data availability scenarios demonstrates the robustness of our sampling strategy recommendation.}
    \label{fig:n_shots}
\end{figure}

Figure~\ref{fig:n_shots} presents results comparing the best-performing approach identified in our main experiments (dataset cycling) against the random sampling baseline under varying levels of data scarcity (N=5, 10, 20, and 40 samples per class). The consistent performance advantage of dataset cycling across all tested sample sizes confirms the robustness of our findings. This suggests that prioritizing pretraining dataset diversity during backbone selection remains an effective strategy regardless of exactly how scarce the training data is, though the absolute performance naturally improves as more samples become available.

\end{document}